\documentclass{article}

\usepackage{PRIMEarxiv}
\usepackage[utf8]{inputenc} % allow utf-8 input
\usepackage[T1]{fontenc}    % use 8-bit T1 fonts
\PassOptionsToPackage{hyphens}{url}  % Pass options to url package before loading it
\usepackage[numbers]{natbib}  % Change this line
\usepackage{hyperref}       % hyperlinks
\usepackage{booktabs}       % professional-quality tables
\usepackage{amsfonts}       % blackboard math symbols
\usepackage{nicefrac}       % compact symbols for 1/2, etc.
\usepackage{microtype}      % microtypography
\usepackage{lipsum}
\usepackage{fancyhdr}       % header
\usepackage{graphicx}       % graphics - LOAD ONCE
\graphicspath{{media/}}     % organize your images and other figures under media/ folder

\usepackage{amssymb}
\usepackage{amsmath}
\usepackage{caption} % Keep this
\usepackage{algorithm}
\usepackage{algorithmic}

\title{DEMENTIA-PLAN: An Agent-Based Framework for Multi-Knowledge Graph Retrieval-Augmented Generation in Dementia Care
}
\author{
Yutong Song$^{1}$, Chenhan Lyu$^{1}$, Pengfei Zhang$^{1}$, Sabine Brunswicker$^{3}$, Nikil Dutt$^{1}$, Amir Rahmani$^{1,2}$\\
\small$^{1}$Donald Bren School of Information and Computer Sciences, University of California Irvine, Irvine, US\\
\small$^{2}$Sue \& Bill Gross School of Nursing, University of California Irvine, Irvine, US\\
\small$^{3}$Purdue University, West Lafayette, US
}

\begin{document}
\maketitle

\begin{abstract}
Mild-stage dementia patients primarily experience two critical symptoms: severe memory loss and emotional instability. To address these challenges, we propose DEMENTIA-PLAN, an innovative retrieval-augmented generation framework that leverages large language models to enhance conversational support. Our model employs a multiple knowledge graph architecture, integrating various dimensional knowledge representations including daily routine graphs and life memory graphs. Through this multi-graph architecture, DEMENTIA-PLAN comprehensively addresses both immediate care needs and facilitates deeper emotional resonance through personal memories, helping stabilize patient mood while providing reliable memory support. Our notable innovation is the self-reflection planning agent, which systematically coordinates knowledge retrieval and semantic integration across multiple knowledge graphs, while scoring retrieved content from daily routine and life memory graphs to dynamically adjust their retrieval weights for optimized response generation. DEMENTIA-PLAN represents a significant advancement in the clinical application of large language models for dementia care, bridging the gap between AI tools and caregivers interventions. 
\end{abstract}

\section{Introduction}

Dementia care poses a significant challenge in modern healthcare systems, affecting millions of patients worldwide \cite{arvanitakis2019diagnosis}. 
Individuals with dementia primarily experience two critical symptoms: severe memory loss and emotional instability. 
Memory loss causes patients to forget recent events, familiar faces, and daily routines, while emotional instability can lead to sudden agitation, anxiety, and confusion in hospital environments \cite{lee2024community}. 
These symptoms often intertwine, as the inability to recall important memories or recognize loved ones frequently triggers emotional distress and behavioral challenges \cite{arvanitakis2019diagnosis}.

Traditional care approaches often struggle to provide consistent and personalized emotional support while managing these behavioral challenges effectively, which often fail to address the nuanced needs of dementia patients. 
Recent studies have shown that leveraging patients' personal histories, shared experiences with family and friends, and familiar social relationships, can significantly improve both cognitive engagement and emotional stability. 
Particularly, approaches incorporating life story books \cite{nyamathi2024establishing, brodaty2009family} have demonstrated that when patients engage with familiar memories and social connections, they experience reduced agitation and improved emotional well-being. 
These personal narratives serve as emotional anchors, helping patients feel more secure and connected to their identity. 
However, implementing such personalized approaches consistently across healthcare settings remains challenging due to limited healthcare resources and the complexity of maintaining individualized care at scale \cite{sadeghian2024methods, rabins2006practical}.

\begin{figure}[t]
\centering
\includegraphics[width=0.5\columnwidth]{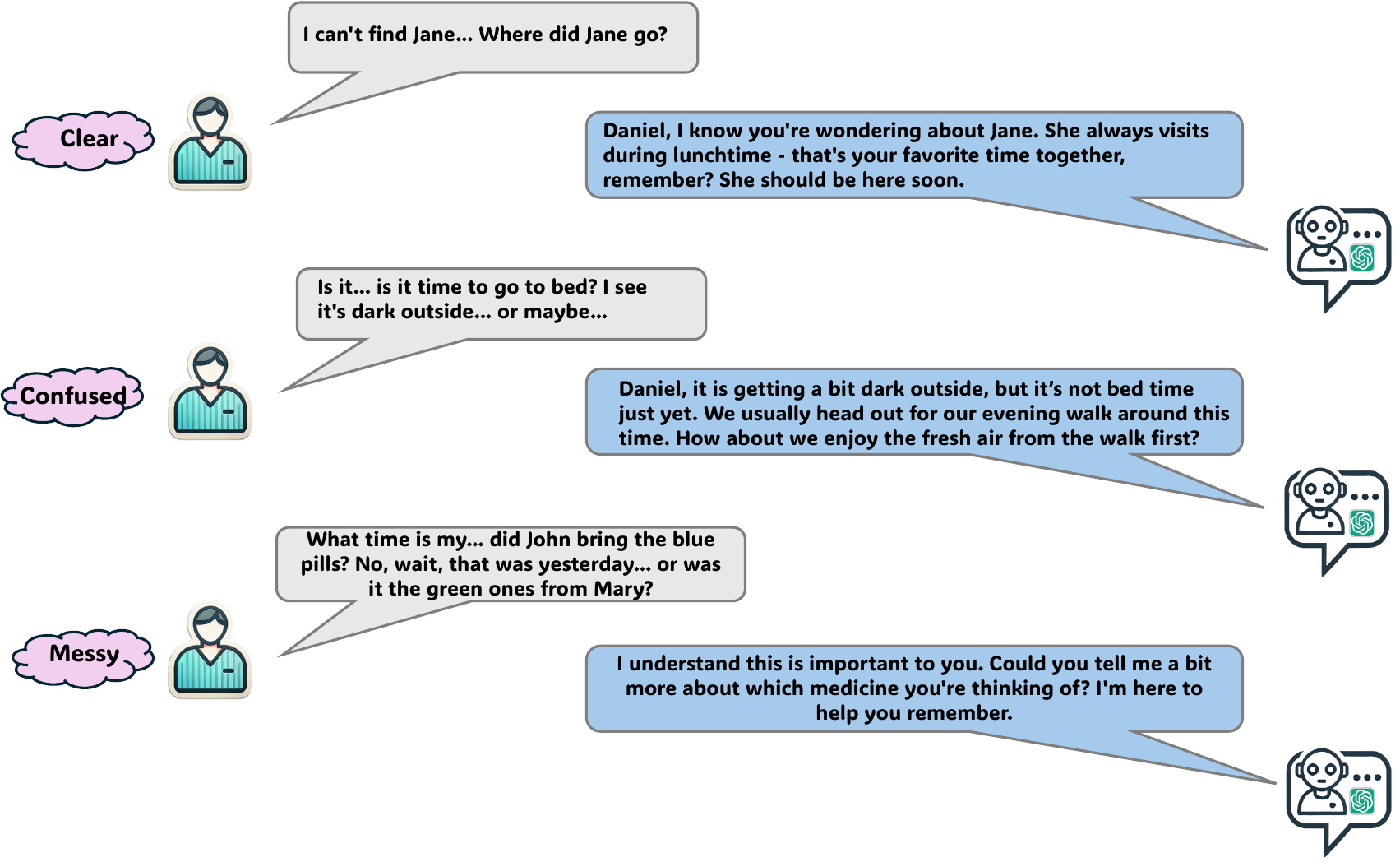} 
\caption{Response examples of DEMENTIA-PLAN for different dialogue types}
\end{figure}
The development of large language models (LLMs) presents a promising solution for dementia care, given their advanced capabilities in natural language understanding and generation. These models excel at comprehending patient queries and generating contextually appropriate responses \cite{treder2024introduction, 10649066,10.1145/3640544.3645230}. However, to effectively assist dementia patients, the system should access and integrate various types of external knowledge about the patient's life and care needs.
Retrieval-augmented generation (RAG) provides a robust framework for augmenting LLMs with external knowledge sources, making it particularly valuable for addressing memory-related challenges in dementia care \cite{edge2024local}. Knowledge graphs (KG) complement this approach by providing structured representations of different aspects of patients' lives \cite{asai2023self, Zhu2024REALMRE}. They are especially effective in capturing complex social relationships and preserving meaningful life stories, which are crucial for maintaining patients' connections to their personal history and loved ones.
 However, effectively integrating information from these diverse KGs presents significant challenges, as each graph focuses on distinct aspects of patient care. Daily routine graphs emphasize temporal patterns and immediate care requirements, while life memory graphs capture long-term personal histories and emotional connections. The heterogeneous nature of these knowledge sources makes it particularly challenging to determine how to optimally combine and prioritize information from different graphs when generating responses. This challenge is further complicated by the need to simultaneously maintain factual accuracy in care instructions while preserving emotional resonance in personal interactions \cite{zhao2024expel}.To address these challenges, we draw inspiration from recent advances in LLM Agent planning, particularly the concept of self-reflection and refinement. Reflection mechanisms have proven crucial in enhancing system robustness through error detection and correction \cite{shinn2024reflexion}, as demonstrated by Self-refine \cite{madaan2024self}, which implements an iterative refinement loop where the LLM generates, evaluates, and refines plans based on self-generated feedback. Building upon this insight, we propose a self-reflection agent that dynamically evaluates and adjusts the integration of information from different KGs. 

In this paper, we propose DEMENTIA-PLAN, a novel multi-KG RAG framework specifically designed for generating LLM responses with dementia patients. To our knowledge, DEMENTIA-PLAN represents the first planning-enhanced RAG framework specifically designed for dementia care that integrates multi-perspective KGs.

%This dual-perspective approach serves different yet complementary purposes - while the daily life graph assists in practical task planning and routine management, the personal narrative graph helps patients reconnect with familiar relationships and positive memories. These rich, personalized information sources enable our LLM to generate more empathetic and context-aware responses. 

\section{Methodology}
\subsection{Framework overview}
% This work presents an empathetic planning agent for dementia patients. The agent integrates daily routines and life memories to provide emotionally resonant and factually accurate responses. It continuously evaluates and enhances its responses through self-reflection, demonstrating persistence in accuracy while being sensitive to patient needs.
DEMENTIA-PLAN presents an innovative dialogue system designed specifically for dementia patients, focusing on generating emotionally resonant and contextually appropriate responses to patient interactions. Examples are shown in Figure~\ref{fig:example}.The system processes patient dialogue inputs, which may include daily needs, emotional expressions, or memory-related concerns, and generates empathetic, personalized responses through a sophisticated multi-KG RAG framework. The system's output is carefully crafted to provide responses that directly address the patient's current needs and concerns, assist in memory recall or schedule organization when appropriate, maintain factual accuracy while ensuring emotional resonance, and employ a gentle and supportive tone to minimize patient anxiety. The framework is illustrated in Figure~\ref{framwork}.

\begin{figure}[ht]
\centering
\includegraphics[width=0.5\columnwidth]{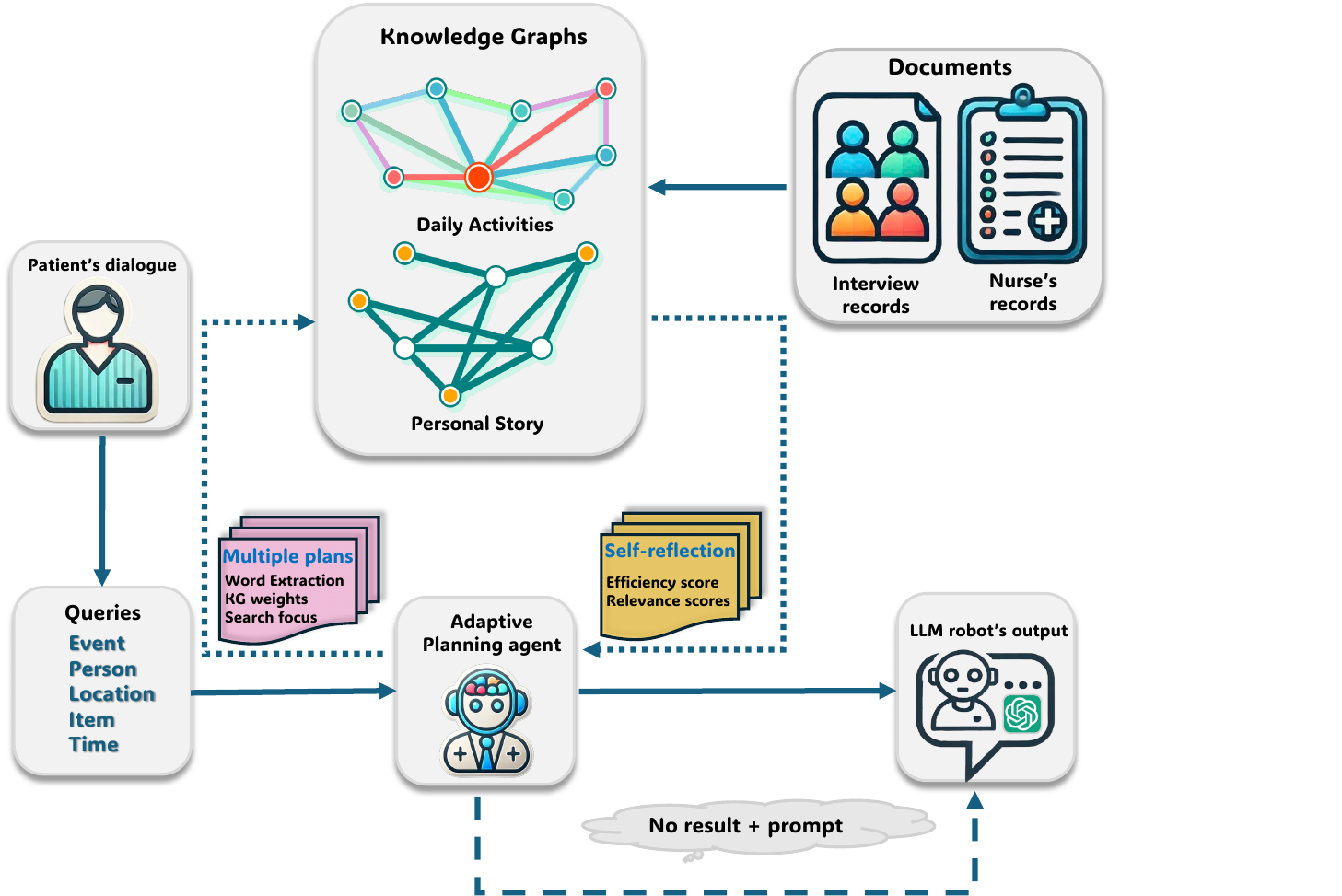} 
\caption{DEMENTIA-PLAN framework}
\end{figure}

 These KGs serve as the foundation for generating responses that are both factually accurate and emotionally supportive. At the heart of DEMENTIA-PLAN is a planning agent powered by LLM, which comprehends the patient's current emotional state and immediate needs, retrieves relevant information from both KGs, generates contextually appropriate responses, and continuously evaluates and refines response quality through self-reflection. What distinguishes DEMENTIA-PLAN is its iterative self-improvement process, where the LLM-based agent continuously evaluates response quality and adjusts its search strategies through graph weight optimization and keyword expansion. This ensures that responses maintain both accuracy and emotional appropriateness over time. The system demonstrates that technical precision and emotional intelligence can be effectively combined in healthcare applications, particularly in supporting the unique needs of dementia patients. The framework implements a dynamic response generation pipeline that begins with patient dialogue input and culminates in emotionally intelligent, contextually relevant outputs. The KGs are continuously updated and optimized based on interaction patterns and response effectiveness, ensuring that the system's support remains personalized and adaptive to the patient's evolving needs.

\subsection{KG Building-up}

%The patient's info are in two aspect: life memory graphs and routine graphs. The life memory graph is built based on the interview from patient's family members and friends. The life story graphs comprise three types of nodes: patients, social connections (family/friends/interviewers), and life events, with edges representing relationships like 'experienced' and 'recalled'. Each node carries specific attributes - person nodes contain demographic information and roles, while event nodes store temporal data, event descriptions, and impact assessments. These graphs capture the longitudinal history of patients' lives, enabling efficient querying of life patterns and social connections that may be relevant to their current care.%
The system's core architecture leverages two specialized KGs: a daily routine graph that maintains information about the patient's schedules and regular activities, and a life memory graph that preserves significant life memories and personal history. The daily routine graph is primarily constructed from objective observations and detailed records maintained by hospital nurses, focusing on precise documentation of events, activities, and care procedures with clinical accuracy. In contrast, the life memory graph is constructed based on interviews conducted with the patient's family members and friends, capturing more subjective content including personal emotional responses..

% The life story graphs consist of three types of nodes: patients, social connections (including family members, friends, and interviewers), and life events. Edges in these graphs represent relationships such as ‘experienced’ and ‘recalled’. Each node carries specific attributes. Person nodes contain demographic information and roles, while event nodes store temporal data, event descriptions, and impact assessments. These graphs effectively capture the patient’s longitudinal life history, facilitating efficient queries of life patterns and social connections that may be pertinent to their current care.
 The memory graphs employ two node types: person as patients social connections (family members, friends), and life events, connected by edges representing 'experienced' relationships. Person nodes contain demographic information and roles, while event nodes store temporal data, descriptions, and impact assessments.
Daily routine graphs focus on care patterns through person nodes (patients and caregivers) and activity nodes (medications, meals, therapy sessions), linked by edges indicating participation types ('participates', 'supervises'). Activity nodes include time slots, locations, and descriptions.

These two graphs serve distinct yet complementary purposes with different temporal characteristics. The daily routine graph, primarily constructed from objective observations and detailed records maintained by hospital nurses, operates on a more immediate time scale and is highly time-sensitive. This temporal precision allows the system to pinpoint a patient's current activities and context based on the timestamp of their dialogue input, enabling more contextually appropriate responses. For example, if a patient expresses confusion during what should be their medication time, the system can provide relevant, time-specific guidance. Figure 3 illustrates a daily routine KG of a patient.

\begin{figure}[t]
\centering
\includegraphics[width=0.4\columnwidth,angle=270]{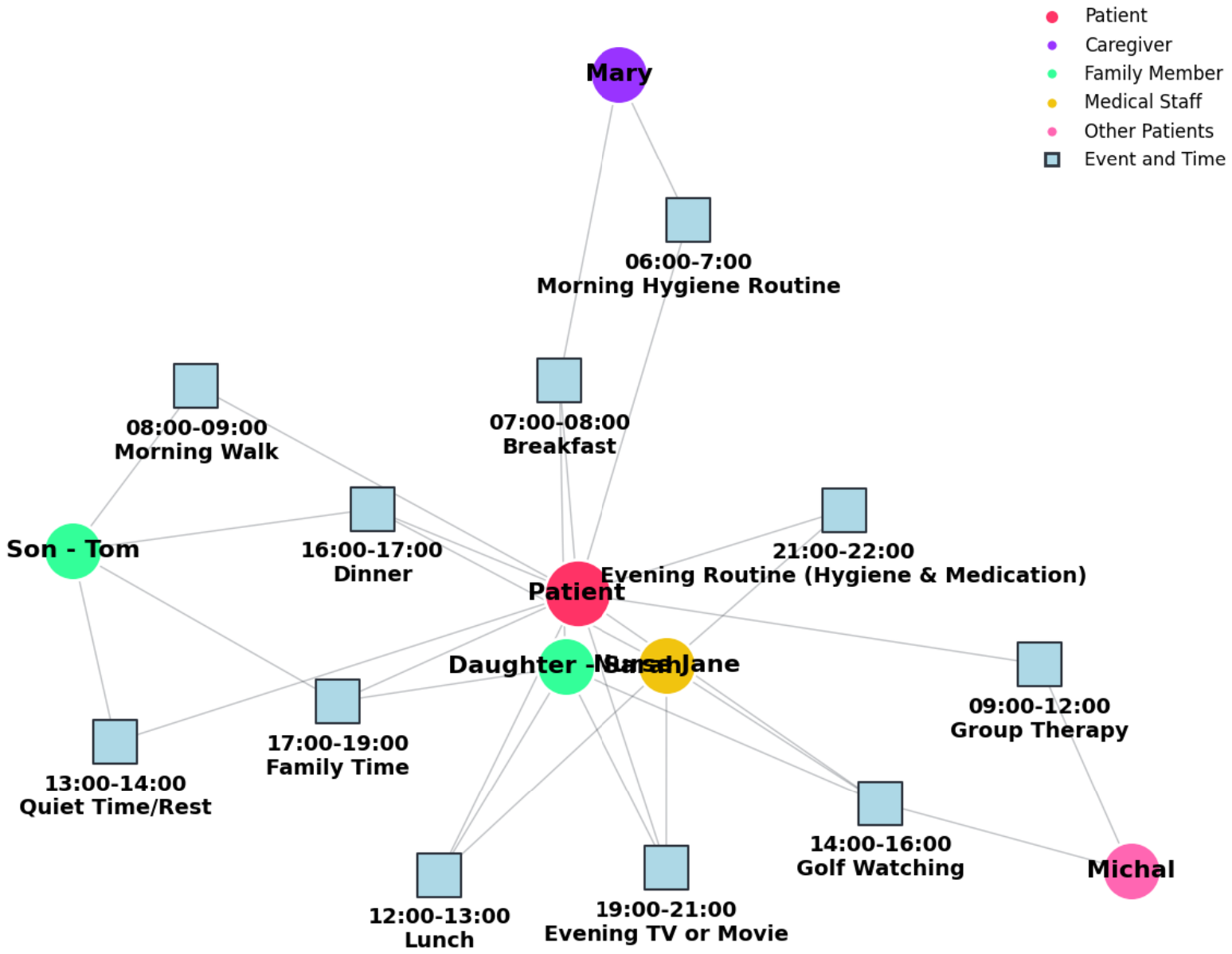} 
\caption{Daily routine KG of a patient}
\end{figure}
\subsection{Adaptive Planning Agent For Graph Retrieval}

To facilitate effective support for dementia patients, we developed a self-reflective planning agent that leverages LLM and multiple KGs. Algorithm 1 shows working steps of the planning agent.

Let $D$ be the user dialogue, $T$ be the current time, and $G_d$, $G_m$ represent the Daily Routine and Memory KGs respectively. The process can be formalized as follows:

\subsubsection{1. Decomposition}
The agent leverages LLM capabilities to analyze dialogue context and decompose queries into distinct semantic categories:
\begin{equation}
Q = \{q_p, q_l, q_i, q_e\} \leftarrow \text{LLM}_{\text{decompose}}(D)
\end{equation}
where $q_p$, $q_l$, $q_i$, $q_e$ represent person, location, item, and event components respectively. This structured decomposition provides the foundation for targeted graph searches, enabling precise retrieval of relevant information based on semantic understanding of the patient's query.

\subsubsection{2. Search Planning}
The search process is executed independently across two KGs, initially weighted equally as $w_d$ and $w_m$.

For current time $T$, we first identify relevant activities and nodes:
\begin{equation}
e_c = \text{FindCurrentActivity}(T)
\end{equation}

In each KG, nodes are scored by combining their semantic relevance to the query with the graph's current weight:
\begin{equation}
score(n) = relevance(n, K) \cdot w_g
\end{equation}
where $w_g$ represents the respective graph weight. The top-k scoring nodes are selected from each graph are noted as $N_d$ and $N_m $.

\subsubsection{3. Evaluation Feedback}
Node selection is determined by a dual-scoring mechanism. The efficiency score $\eta$ is computed as:
\begin{equation}
\eta = \text{LLM}_{\text{evaluate}}(N, D)
\end{equation}
where $N = \{e_c\} \cup N_d \cup N_m$ represents the combined relevant nodes. This evaluation assesses how well the retrieved nodes address the patient's dialogue context, considering both factual accuracy and contextual appropriateness.

\subsubsection{4. Adaptive Learning and Iterative Optimization}
When the efficiency score falls below a predetermined threshold $\theta$, the agent initiates its adaptive learning mechanism:
\begin{equation}
\{w_d', w_m'\} = \text{reflection}_{\text{adjust}}(w_d, w_m)
\end{equation}
The search scope is progressively refined through keyword expansion:
\begin{equation}
K = K \cup \text{reflection}_{\text{suggest}}(K)
\end{equation}
This iterative process optimizes the search strategy by:
1) Dynamically adjusting the weights of both KGs based on their demonstrated utility.
2) Expanding primary search terms into related concepts for comprehensive coverage.

The process continues until either sufficient nodes are found ($\eta \geq \theta$) or maximum attempts $M$ are reached:
\begin{equation}
R = \begin{cases}
\text{LLM}_{\text{generate}}(N) & \text{if } \eta \geq \theta \\
\text{LLM}_{\text{prompt}}() & \text{otherwise}
\end{cases}
\end{equation}
This self-reflection process can repeat up to $max_attepmt$ times. If the response does not satisfy after that, the agent generates a follow-up prompt as $LLM.prompot$ to request more specific information from the patient, ensuring meaningful and context-aware interactions.

\begin{algorithm}
\footnotesize
\caption{Self-reflection planning agent procedure}
\textbf{Input}: User dialogue and current time\\
\textbf{Parameter}: max\_attempt, threshold\\
\textbf{Output}: Response or follow-up prompt
\begin{algorithmic}
    \STATE keywords $\gets$ ExtractKeywords(dialogue)
    \STATE count $\gets$ 0
    \STATE weight\_daily, weight\_memory $\gets$ 0.5, 0.5
    \WHILE{count $<$ max\_attempt}
        \STATE current\_event $\gets$ FindCurrentActivity(time)
        \STATE event\_nodes $\gets$ SearchDailyKG $*$ weight\_daily
        \STATE memory\_nodes $\gets$ SearchMemoryKG $*$ weight\_memory
        \STATE nodes $\gets$ \{current\_event, event\_nodes, memory\_nodes\}
        \STATE reflection $\gets$ LLM.evaluate(nodes, dialogue)
        \IF{efficiency\_score $>$ threshold}
            \STATE \textbf{return} LLM.generate(nodes)
        \ENDIF
        \STATE weight\_daily, weight\_memory $\gets$ reflection.adjust\_weights()
        \STATE keywords $\gets$ keywords $\cup$ reflection.suggest\_keywords()
        \STATE count $\gets$ count + 1
    \ENDWHILE
\STATE \textbf{return} LLM.prompt()
\end{algorithmic}
\end{algorithm}

\section{Synthetic Dementia-Patient Dataset}
\begin{table*}[ht]
\centering
\renewcommand{\arraystretch}{1.2} % Reduce row height
\setlength{\tabcolsep}{4pt} % Reduce column padding
\small % Reduce font size for the table
\begin{tabular}{|p{5cm}|p{10cm}|}
\hline
\quad\quad \quad\quad  \textbf{Confusion Type} & \quad \quad \quad \quad\quad\quad\quad\quad\quad \quad\quad\quad\quad \textbf{Example} \\
\hline
Confuse Past and Present Events & "Did Tom come by earlier? I thought I heard him talking in the kitchen." \\
\hline
Misremember Scheduled Activities & "Don’t we have the gardening club this morning? I should get my gloves ready." \\
\hline
Reference Non-Existent Appointments & "Will Sarah visit this afternoon? She usually comes around this time, doesn’t she?" \\
\hline
Confusion About current dates & "Isn’t it Christmas soon? I need to get the decorations out." \\
\hline
Reference Incorrect Location/Context & "Where are we right now? This doesn’t look like my house." \\
\hline
Repeat Questions  & "When is lunch? Did you say it’s at 12? Or is it later?" \\
\hline
Confusion About Life Stage  & "I have to prepare for school tomorrow. Have we packed my bag yet?" \\
\hline
Incomplete or Vague Statements & "That man… um… he said he would come today, didn’t he?" \\
\hline
Environmental Confusion & "This garden feels familiar. Did we plant these flowers together last year?" \\
\hline
\end{tabular}
\caption{Types of Confusion and Examples}
\label{table:confusion_examples_longer}
\end{table*}

We developed DementiaGraph, a synthetic dataset generated through the GPT-4-mini model \cite{openai2023gpt4} , comprising detailed documentation of 100 dementia patients. The dataset integrates three key components: individual patient profiles, daily living activity logs, and structured interview summaries capturing their social connections and relationship networks.

The daily activity data was derived from nursing records documenting patients' complete 24-hour routines in the care facility. Each record captures various daily events, including morning exercises, meal times, and family visits. The documentation for each event includes specific timestamps, participating individuals, and detailed descriptions of the activities performed. 

For patient dialogues, we maintained a consistent generation approach while incorporating specific confusion patterns characteristic of dementia patients, which are derived from clinical research. These patterns primarily manifest as information gaps and factual inconsistencies, reflecting the memory impairment typical of dementia. For example, patients may exhibit incomplete or inaccurate recall of events, demonstrating patterns of confusion that align with documented clinical observations. our dementia dialogue set is consist of 80\% clear questions, that are no confusion or incomplete part, and 20\% confused questions, that appear with some mistakes. All confusion types are shown in Table 1.

% \begin{table*}[h]
% \begin{tabular}{|p{0.35\textwidth}|p{0.65\textwidth}|}
% \hline
% \textbf{Type of Confusion} & \textbf{Example} \\
% \hline
% Confuse Past and Present Events  & "Did Tom come by earlier? I thought I heard him talking in the kitchen." \\
% \hline
% Routine Memory Impairment & "Don't we have the gardening club this morning? I should get my gloves ready." \\
% \hline
% False Memory Creation & "Will Sarah visit this afternoon? She usually comes around this time, doesn't she?" \\
% \hline
% Confusion about current dates & "Isn't it Christmas soon? I need to get the decorations out." \\
% \hline
% Spatial/Environmental Confusion & "Where are we right now? This doesn't look like my house." \\
% \hline
% Repeat Questions  & "When is lunch? Did you say it's at 12? Or is it later?" \\
% \hline
% Age/Life Stage Confusion & "I have to prepare for school tomorrow. Have we packed my bag yet?" \\
% \hline
% Communication Fragmentation & "That man... um... he said he would come today, didn't he?" \\
% \hline
% Familiar Setting Recognition Issues & "This garden feels familiar. Did we plant these flowers together last year?" \\
% \hline
% \end{tabular}
% \caption{Types of Cognitive Confusion in Dementia with Representative Examples}
% \end{table*}

\section{Experiments}

In this study, we conducted a comprehensive evaluation of our model using gold test data carefully curated by domain specialists. The evaluation framework encompasses both automated metrics and LLM-based judgment to ensure a thorough assessment of model performance. This multi-faceted evaluation approach allows us to capture both quantitative performance metrics and qualitative aspects of model behavior, providing a more holistic assessment of the system's capabilities.

\subsection{Automated Metrics}
For automated evaluation, we employed two widely-recognized NLP metrics. First, we utilized ROUGE scores (Recall-Oriented Understudy for Gisting Evaluation), which measure the lexical overlap between model outputs and reference texts. This includes ROUGE-1 for unigram overlap and ROUGE-1 for bigram matching. Additionally, we incorporated BERTScore, a more semantically-aware metric that leverages contextual embeddings from BERT to capture deeper linguistic similarities beyond surface-level word matching. The result are shown in Table 2.

\subsection{LLM-based Evaluation}

To systematically evaluate our proposed DEMENTIA-PLAN framework, we developed a comprehensive assessment methodology comparing three model variants against human performance (Gold standard). The evaluation was conducted through our LLM-based framework across five critical dimensions of dementia care support.

Our comparative analysis included:
\begin{itemize}
    \item \textbf{Baseline 1:} A baseline system equipped with only the Daily Routine KG
    \item \textbf{Baseline 2:} An enhanced version incorporating both Daily Routine and Memory KGs, but without the planning agent
    \item \textbf{Full Model:} Our complete DEMENTIA-PLAN framework featuring both KGs and the planning agent for iterative response refinement
    \item \textbf{Gold Standard:} Human caregiver responses serving as the performance benchmark
\end{itemize}

\subsubsection{Evaluation Dimensions}
The assessment framework evaluates five key aspects of model performance:

\begin{itemize}
    \item \textbf{Coherence:} Evaluating the model's ability to maintain simple and logical communication with dementia patients, adapting to topic shifts and maintaining clear conversational threads despite potential confusion or disorientation.
    
    \item \textbf{Empathy:} Assessing the model's capability to understand and appropriately respond to emotional states, demonstrate patience with repeated questions, and maintain a consistent, warm, and understanding tone throughout interactions.
    
    \item \textbf{Memory Support:} Evaluating the model's effectiveness in assisting with daily memory tasks, recognizing individual memory patterns, and providing appropriate memory cues and gentle reminders for essential activities.
    
    \item \textbf{Emotional Safety:} Evaluating the model's capability to maintain a secure conversational environment by avoiding sensitive triggers, navigating around topics that may cause agitation, and providing gentle reassurance during moments of confusion without invalidating patient experiences.
    
    \item \textbf{Problem Solving:} Assessing the model's capability in providing clear, step-by-step guidance for daily activities, offering practical solutions to common challenges, and adapting support based on individual patient needs.
\end{itemize}

This comprehensive evaluation framework enables us to quantitatively assess each model's performance while maintaining focus on the critical aspects of dementia patient care and support. We use GPT-4o-mini model as the judge \cite{openai2023gpt4}.

As shown in Table 3, we evaluated each model's performance using a 0-10 scale across all dimensions. To facilitate direct comparison with human performance, we normalized all scores relative to the Gold standard (human responses), treating the Gold scores as the maximum benchmark (10 points). This normalization is visualized in Figure~\ref{fig:radar-evaluation} through a radar chart.
The radar visualization clearly demonstrates the progressive improvement from Baseline 1 to our Full model, with the Full model's performance profile most closely approximating the Gold standard. This comparative analysis not only validates the contribution of each component in our framework but also highlights the synergistic effects achieved through their integration in the DEMENTIA-PLAN system.

\subsection{Results analysis}

The evaluation results show promising performance in our model's response accuracy, with notably higher scores in Clear Dialogue (ROUGE-1: 0.48, BERTScore: 0.91) compared to Confused Dialogue (ROUGE-1: 0.42, BERTScore: 0.88). The strong performance in clear dialogue scenarios demonstrates the model's effectiveness in providing accurate information and guidance about daily routines. The relatively lower scores in confused dialogue scenarios can be attributed to the model's emphasis on empathetic responses and encouragement when facing unclear or incomplete queries, which introduces higher vocabulary variability as emotional support can be expressed in multiple valid ways. This pattern reflects real-world challenges in dementia care, where healthcare providers often need to balance between providing accurate information and maintaining supportive communication when faced with unclear patient expressions. While our model maintains high semantic understanding (as shown by the BERTScore of 0.88 even in confused scenarios), the lower ROUGE-1 scores indicate potential areas for improvement in handling ambiguous or incomplete queries.

\begin{table}[h]
\centering
\begin{tabular}{lcc}
\hline
                                    \textbf{Dialogue Type } &  \textbf{ROUGE-1} &  \textbf{BERTScore} \\
\hline
                                   Clear Dialogue & 0.48 &   0.91 \\
                                Confused Dialogue& 0.42 &   0.88 \\
                                Overall &0.46 & 0.89\\
\hline
\end{tabular}
\caption{Evaluation for different Dialogue Patterns}
\label{table:Bertscore}
\end{table}

The incorporation of the Memory KG leads to substantial improvements across all metrics, most notably in empathy (7.56→8.40), memory (6.03→7.12), and safety (8.71→9.22). The significant increase in empathy scores aligns with clinical observations in dementia care, where familiar relationships and personal memories tend to reduce patients' emotional agitation. This improvement validates medical findings that dementia patients typically exhibit more stable emotional states when interacting within familiar contexts. The enhanced safety scores (8.71→9.22) challenge the common misconception that discussing personal history might distress patients; instead, it demonstrates that family relationships and life experiences serve as safe and constructive conversational topics. The marked improvement in memory performance (6.03→7.12) highlights the effectiveness of our Memory KG, which systematically records and retrieves key life events of patients. The system demonstrates precise information retrieval capabilities, effectively coordinating between both KGs (Memory KG and Routine KG) to match relevant information with ongoing conversational contexts. These improvements collectively demonstrate that systematically incorporating patients' personal histories not only technically enhances the model's performance but also provides substantial therapeutic value in dementia care.

\begin{table*}[htbp]
\centering
\small 
\begin{tabular}{lcccccccc}
\hline
 &Routine KG & Memory KG & Planning Agent & \textbf{Coherence} & \textbf{ Empathy } & \textbf{Memory} & \textbf{Safety} & \textbf{Problem Solving} \\
\hline
Baseline 1 &  \checkmark  &$\times$& $\times$  & 8.12 & 7.56 & 6.03 & 8.71 & 7.95 \\
Baseline 2& \checkmark& \checkmark & $\times$ & 8.76 & 8.40 & 7.12 & 9.22 & 8.01 \\
Full & \checkmark & \checkmark & \checkmark & 8.95 & 8.48 & 7.88 & 9.44 & 8.60 \\
Gold & - &  - & - & 9.46 & 8.48 & 8.06 & 9.49 & 9.32 \\
\hline
\end{tabular}
\caption{Comparison of Gold and Model Scores Across Domains (0-10)}
\end{table*}

\begin{figure}[ht]
\centering
\includegraphics[width=0.4\columnwidth,angle= 270]{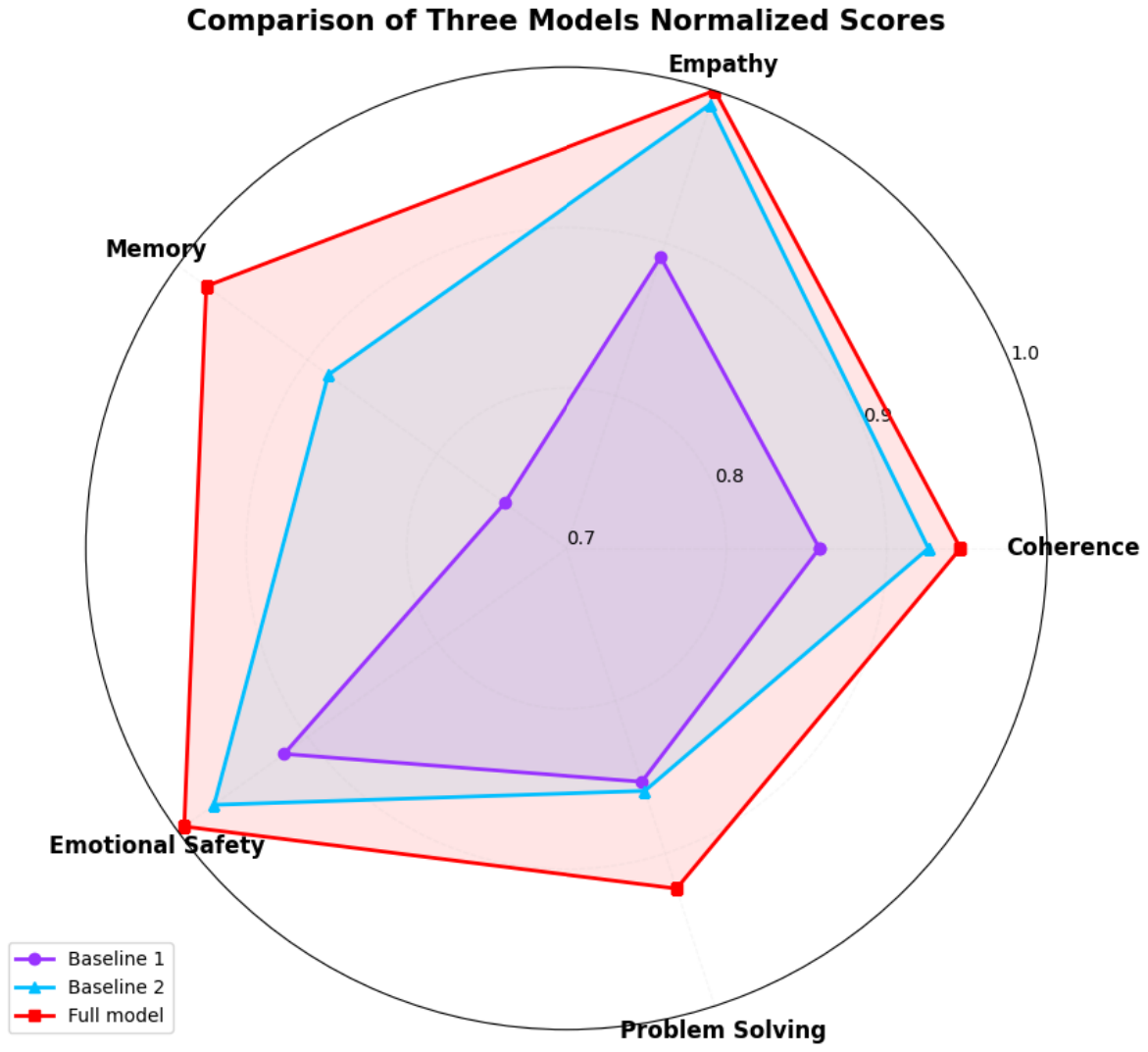} 
\caption{Radar visualization of model performance across evaluation dimensions, normalized against human (Gold) standard responses}
\end{figure}

The addition of the Planning Agent to create the Full model demonstrates significant improvements, particularly in problem-solving capabilities (8.01→8.60). This enhancement validates our agent's superior ability to extract and integrate information from both KGs (Memory KG and Routine KG) to address patients' concerns, whether they relate to memory loss about past events or confusion about current activities. The agent's effectiveness is evident in two key aspects: first, in bridging historical knowledge with present needs, and second, in formulating structured solutions that account for both the patient's background and current situation. The improvement in coherence scores (8.76→8.95) further demonstrates how the Planning Agent contributes to more organized and systematic interactions, especially crucial when providing multi-step guidance for daily activities. This structured approach is particularly valuable for dementia patients who often struggle with complex task sequences and require clear, well-organized instructions. The synergy between the KGs and planning capabilities creates a more comprehensive support system that can effectively address both memory-related challenges and daily activity management.

\subsection{Limitations}
The Full model demonstrates both promising achievements and remaining challenges when compared to human performance. Notably, it matches the Gold standard in empathy (both 8.48), indicating successful implementation of emotional support through our KG and planning approach. However, gaps remain in problem-solving (8.60 vs 9.32) and coherence (8.95 vs 9.46), where human caregivers excel in handling complex situations and maintaining natural conversation flow, particularly during disorganized patient interactions. The smaller differences in memory support (7.88 vs 8.06) and safety (9.44 vs 9.49) suggest that while our system effectively manages basic care interactions, human caregivers still demonstrate superior adaptability in sophisticated memory support and sensitive situation handling.

\section{Discussion and Future Work}

While our current framework shows promise in single-turn empathetic conversations, several directions for future research could enhance its capabilities. We plan to extend DEMENTIA-PLAN to support dynamic multi-turn conversations, where the system can not only utilize existing knowledge but also adaptively refine search strategies and update KGs based on patient feedback during interactions. Additionally, recognizing that dementia care requires a holistic approach, we aim to integrate medical KGs that capture patients' psychological and physical health states, enabling our system to generate responses that consider their overall well-being alongside their personal histories and daily routines. These enhancements would transform DEMENTIA-PLAN into a more comprehensive care assistance system that adapts and grows with patient interactions while maintaining emotional safety and empathetic engagement.

\section{Conclusion}

In this paper, we presented DEMENTIA-PLAN, a novel framework for supporting dementia patients through the integration of multiple KGs and Large Language Models. Our main contributions include a multi-KG architecture that combines daily routines and life memories, a self-reflection planning agent that dynamically optimizes information retrieval through continuous evaluation and weight adjustment, and experimental validation demonstrating significant improvements in both retrieval accuracy and emotional intelligence of responses. Looking ahead, we plan to extend DEMENTIA-PLAN to support dynamic multi-turn conversations and integrate additional medical KGs, while developing more sophisticated adaptation mechanisms that can refine both search strategies and KG content based on ongoing patient interactions. These innovations and future directions represent meaningful steps toward more empathetic and effective AI-assisted healthcare solutions in dementia care.

\section*{Acknowledgements}
Funding for this research was provided by the Noyce Institute. We express our sincere gratitude for their generous support, which made this work possible.

\bibliographystyle{unsrt}  
\bibliography{main}  

\end{document}